\documentclass[runningheads]{llncs}
\usepackage[T1]{fontenc}
\usepackage{graphicx}
\usepackage{hyperref} 

\makeatletter
\newcommand{\printfnsymbol}[1]{%
  \textsuperscript{\@fnsymbol{#1}}%
}
\makeatother

\usepackage{color}

\begin{document}
\title{Alt4Blind: A User Interface to Simplify Charts Alt-Text Creation}
\titlerunning{Alt4Blind}


\author{Omar Moured\inst{1,2}\thanks{Both authors contributed equally to this work.}
\and
Shahid Ali Farooqui\inst{3}\printfnsymbol{1} 
\and
Karin Müller\inst{2}
\and
Sharifeh Fadaeijouybari\inst{4}
\and
Thorsten Schwarz\inst{2}
\and
Mohammed Javed\inst{3}
\and
Rainer Stiefelhagen\inst{1,2}
}

\authorrunning{O. Moured and S.A. Farooqi}

\institute{
CV:HCI@KIT, Karlsruhe Institute of Technology, Germany\\
\url{https://cvhci.anthropomatik.kit.edu/}
\and
ACCESS@KIT, Karlsruhe Institute of Technology, Germany\\
\url{https://www.access.kit.edu/}
\and
Computer Vision \& Biometrics Lab, Indian Institute of Information Technology Allahabad, India\\
\url{https://cvbl.iiita.ac.in/}
\and
University of Trier, Germany
}

\maketitle
\begin{abstract}
Alternative Texts (Alt-Text) for chart images are essential for making graphics accessible to people with blindness and visual impairments. Traditionally, Alt-Text is manually written by authors but often encounters issues such as oversimplification or complication. Recent trends have seen the use of AI for Alt-Text generation. However, existing models are susceptible to producing inaccurate or misleading information. We address this challenge by retrieving high-quality alt-texts from similar chart images, serving as a reference for the user when creating alt-texts. Our three contributions are as follows: (1) we introduce a new benchmark comprising 5,000 real images with semantically labeled high-quality Alt-Texts, collected from Human Computer Interaction venues. (2) We developed a deep learning-based model to rank and retrieve similar chart images that share the same visual and textual semantics. (3) We designed a user interface (UI) to facilitate the alt-text creation process. Our preliminary interviews and investigations highlight the usability of our UI. For the dataset and further details, please refer to our project page: \url{https://moured.github.io/alt4blind/.}

\end{abstract}

\section{Introduction}
Making charts accessible to a broader audience is essential to ensure equal access. Part of this audience includes people with blindness and visual impairments individuals who access data through different modalities: tactile and auditory. The choice between these modalities often depends on the available technologies and the preferences of people with blindness and visual impairments. Although technologies like sonification and tactile materials are available, \textit{Alt-Text} remains the primary method for BVI individuals to acquire information about images through screen readers \cite{jeong2023wataa}. In accordance by the WCAG~\footnote{\url{https://www.w3.org/WAI/alt/}}, an alt text is required for a graphic to be accessible.
These summaries are embedded as hidden tags within documents or web pages. A study \cite{disability2004web} revealed that over 80\% of web pages often neglect to include alt text. Furthermore, even when available, it often doesn't comply with recommended guidelines (e.g. WCAG) and standards (e.g. W3C).

Chart alt-text can originate from two primary sources: manually created by humans or (semi-)automatically generated by AI systems. Human-generated descriptions are often accurate but typically require expertise. Recent studies have highlighted a significant Alt-text deficiency in publications \cite{chintalapati2022dataset}. On the other hand, automatically generated captions are quicker to obtain, and require no expertise but may suffer from information inaccuracy. Thus, crafting a high-quality chart alt-text is not a trivial task \cite{lundgard2021accessible}. 

\begin{figure}[h]
    \centering
    \includegraphics[width=\textwidth]{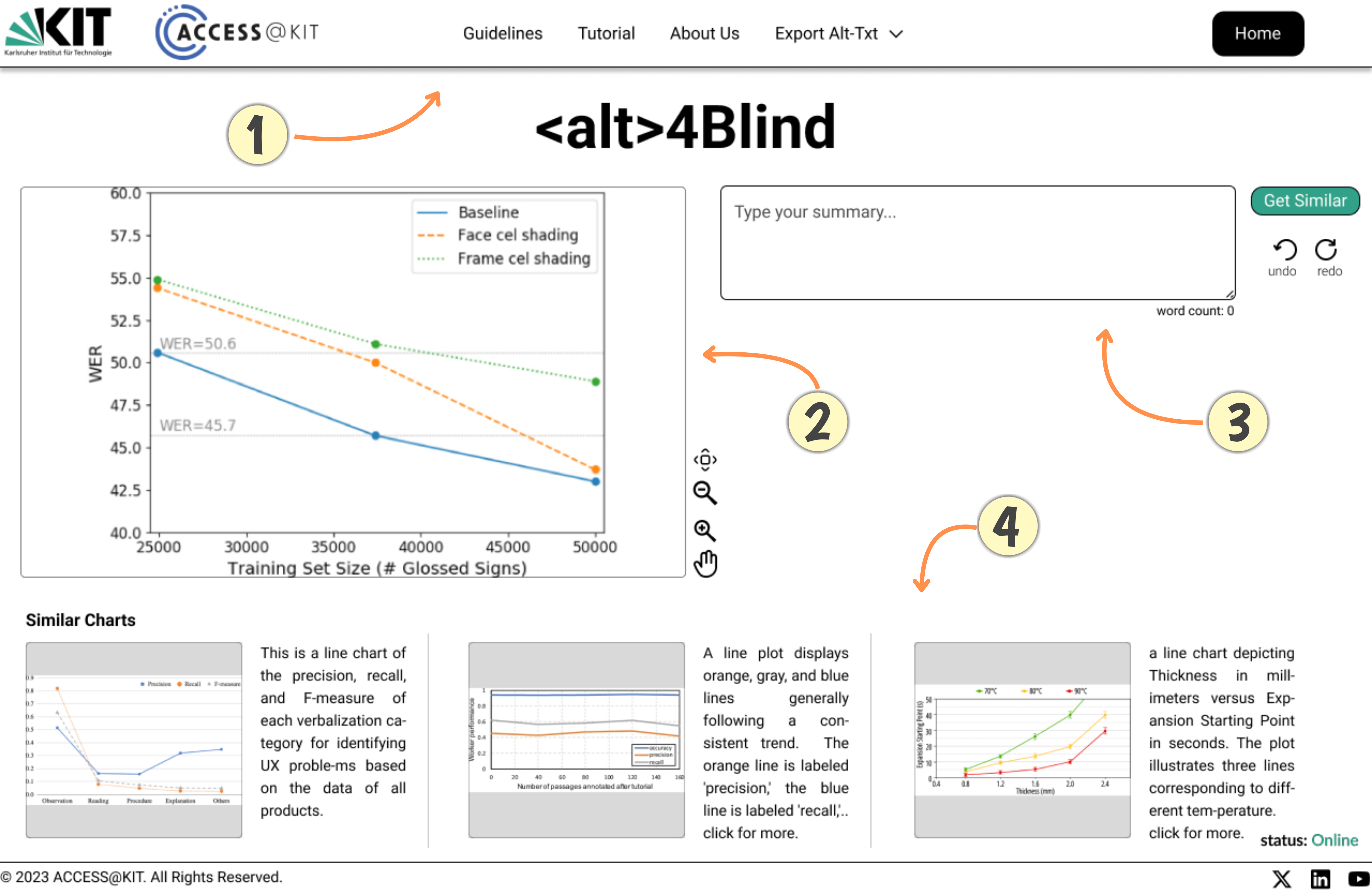}
    \caption{\textit{Alt4Blind} UI: (1) Menu bar offering access to guidelines and a tutorial. (2) Space for uploaded images featuring a function bar (zoom, move, fit). (3) Text field for user input, accompanied by a button to update the retrieved image. (4) Retrieved charts based on the uploaded image, can be further enhanced with text query.}
    \label{fig:main}
\end{figure}

Instead, this work investigates the role of using AI in aiding both experts and non-expert users to write high-quality alt-text by presenting high-quality, similar charts with alt-text as references for them. To achieve this, we built a comprehensive dataset composing 5,000 chart images sourced from venues providing alt-text with their publications. We then trained an AI model to categorize images based on their visual and textual similarities.
Subsequently, the most similar images are retrieved and presented to the user through our \textit{Alt4Blind} interface (Figure \ref{fig:main}). 

\section{Related Work}
In this section, we first examine existing datasets in the field of chart analysis, followed by a discussion on the available models and tools for creating textual descriptions of charts.

\subsection{Datasets}
Chart-2-Text \cite{kantharaj2022chart} is a popular dataset in computer vision. It comprises a large number of chart images of various types, with each image accompanied by summary and bounding box annotations. However, it is not tailored for people with blindness and visual impairments, the summaries are not structured and do not adhere to the accessibility guidelines. In contrast, Vistext \cite{tang2023vistext} comprises 8822 synthesized images, each manually assigned alt-texts, representing a significant effort. However, it features synthetic charts, restricted to line and bar charts, and the alt-texts were not authored by the original image creators, potentially omitting crucial context. Another dataset, HCI Alt-Text \cite{chintalapati2022dataset}, collected chart figures from HCI conferences and aims to analyse the quality of provided alt-texts. However, it consists of 511 chart images, making it less suitable for deep-learning models.

\subsection{Alt-Text Creation Models}
Chart Analysis is an emerging field in computer vision, featuring recent works such as ChartAssistant \cite{meng2024chartassisstant} and ChartLLama \cite{han2023chartllama}. These models are trained for various chart-related tasks, including description generation, chart2table, and chart2code, among others. Despite being trained on a vast chart corpus, these models fall short of addressing accessibility concerns.

\subsection{Tools}
K. Mack \cite{mack2021designing} analyzed various interface designs to facilitate the creation of high-quality alt-texts. Their results concluded that \textbf{interfaces assisting authors in deciding what to include in the alt-text were well-received and led to higher-quality descriptions}. However, there are very few tools that adopt this approach. WATAA \cite{jeong2023wataa} is one such tool, which presents users with AI-generated descriptions as a reference. Nevertheless, AI descriptions are prone to hallucinations.

\paragraph{In this paper, we focus on addressing both limitations: (i) creating a database with high-quality alt-texts, and (ii) developing an intelligent tool that aids people with and without accessibility knowledge to author high-quality alt-text.}
\section{Exploratory Interviews and Findings}
With the help of the Center for Digital Accessibility and Assistive Technologies in Karlsruhe Institute of Technologies~\footnote{\url{www.access.kit.edu}}, we recruited six participants  with varying levels of accessibility expertise, from none to advanced (working in the field of accessibility). We conducted exploratory interviews with these participants (experts (P1-P3), non-experts (P4-P6)) to identify necessary features for the development of an intuitive user interface, the \textit{Alt4Blind} app. 
Participants were not compensated. 

\subsection{Procedure}
The interviews took place in a one-hour face-to-face session following the steps: 
\begin{enumerate}
    \item Initially, we introduced a basic line chart featuring three distinct lines, see the illustration in the upper left of Figure \ref{fig:main}-2.
    
    \item Participants were then instructed to write a description for a blind person.
    
    \item Following this, we inquired about the challenges faced and their perspective on essential features for a web tool designed for this task.
\end{enumerate}

No compensation was offered for participation in this study. Furthermore, the user study received approval from the KIT Internal Ethical Committee.

\subsection{Qualitative Results}
During the interview, we took notes that we analyzed after the sessions. The analysis revealed the following observations:
P1-P3, maintained a coherent reading order, starting with visual semantics (e.g., chart type, title, and axis labels) followed by contextual information (e.g., line trends, and comparisons). In response to the follow-up questions, P1 noted that they usually find Multivariate and Panel charts challenging. P4-P6 discussed how they were unfamiliar with alt-texts as it's often hidden in tags, and suggested the need for an initial guideline before start writing. Based on these insights, we identified as a feature that an initial suggestion for user-uploaded images is necessary. Thus, we chose to \textbf{(1) display a real initial suggestion}, i.e. displaying real chart images with human-authored high-quality alt-texts. \textbf{(2) the images should be visually or contextually relevant to the user chart.}
\newpage

\begin{figure}
    \centering
    \includegraphics[width=\textwidth]{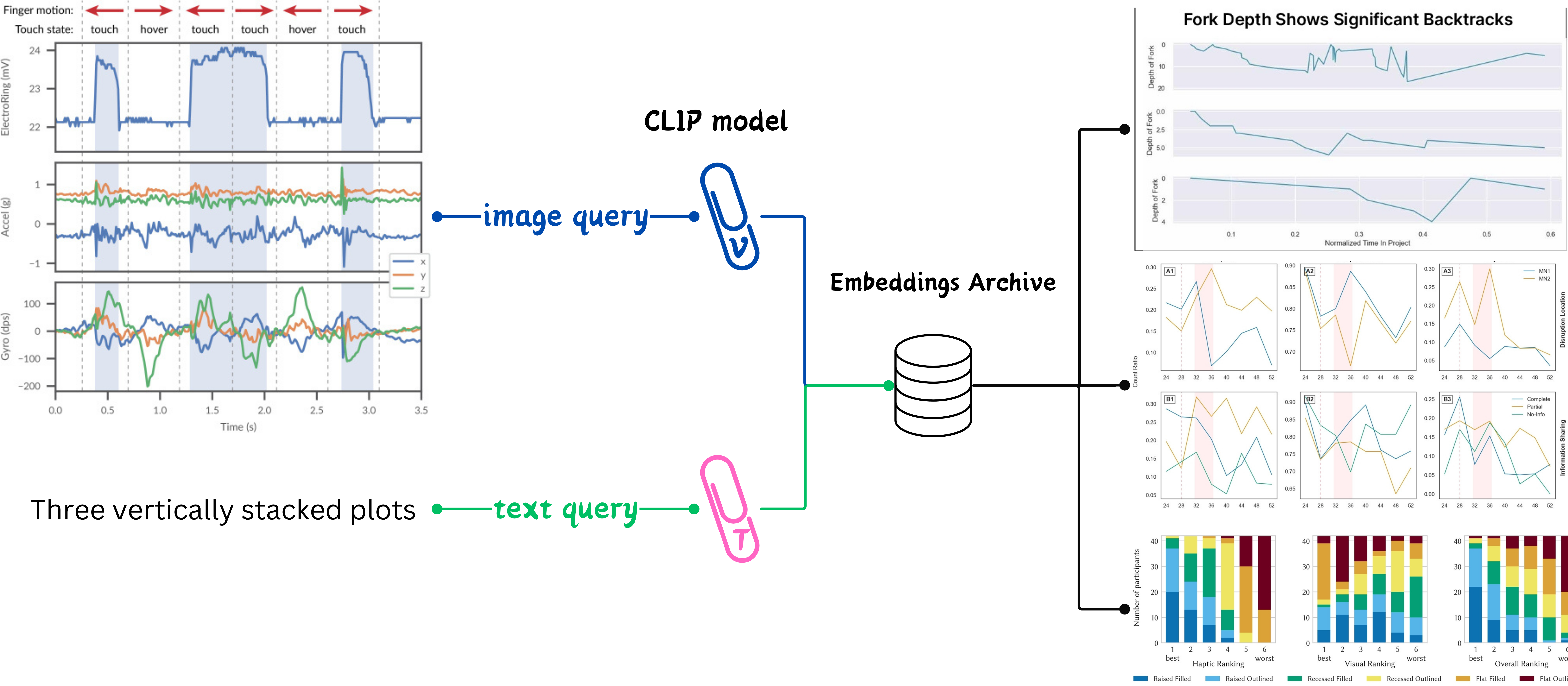}
    \caption{Our retrieval system leverages both the text and image encoder modules of the fine-tuned CLIP model. This ensures similarity at both visual and contextual levels.}
    \label{fig:system}
\end{figure}

\section{Alt4Blind}
In this section, we describe (i) our dataset, (ii) the AI-based model for ranking and retrieving similar charts with alt-texts, and (iii) the creation of the Alt4Blind User Interface.

\subsection{Creation of our Dataset}
We first crawled 25,000 images with alt-text from publicly available HCI-related conferences spanning the last 10 years to expand on the HCI Alt-Text dataset~\cite{chintalapati2022dataset}. In the next step, we filtered out images that are not charts, have a short alt-text or lack semantics, so we end up with 5,000 chart images, marking a tenfold increase from the previous dataset.

Our dataset contains a wide variety of chart types, from familiar forms like Line, Bar, Area, Pie, and Scatter charts to more unique visualizations not present in previous works such as multi-variate and panel charts. We used the 4-level semantic model by Lundgard et al. \cite{lundgard2021accessible} to rate alt-text quality. Initial scoring was conducted using ChatGPT-4 by assessing the number of semantics and levels present in each alt-text. The richest samples were subsequently manually reviewed.

\subsection{Chart Image Retrieval}
Image retrieval involves searching for images similar to a given query image or text, with a focus on ensuring the top results include similar chart semantics. For this purpose, we employed CLIP ViT-B/32 \cite{radford2021learning} as our baseline model. The CLIP architecture encompasses both a text and an image encoder, each producing a 512-dimensional feature vector. The CLIP model uses the unsupervised contrastive pre-training approach to cluster the samples in the latent space. Hence, we fine-tuned the model on our dataset, with a batch size of 16 for 10 epochs. Next, during the inference phase, Figure \ref{fig:system}, we input the image, extract the encoders' representations and compare it to the samples from the datasets. The top three candidates with the highest cosine similarity scores are then retrieved. Our model achieved 92\% in P@3 and 85\% in R@3, demonstrating high capabilities in Precision (P@3) and Recall (R@3) for displaying similar chart images within the top three results, respectively.

\subsection{User Interface}
For our prototype, we utilized React JS\footnote{\href{https://react.dev/}{https://react.dev/}}, a JavaScript library ideal for building user interfaces. It includes a landing page for users to drag-and-drop chart images and a tutorial for first-time use. Once an image is uploaded, the backend model display the top three similar candidates on the main page (see Figure \ref{fig:main}). Users can enlarge to view the full alt-text. The "Get Similar" button allows users to refresh the candidates when they type their summary in the text field. 
\section{Preliminary Evaluation}
We conducted tests with the same participants, P1-P6, to assess our user interface. Participants were presented with Panel and Multivariate charts, which were unfamiliar to all. The samples are depicted in Figure \ref{fig:main}, and participants were instructed to:

\begin{enumerate}
    \item Review the guidelines and participate in the tutorial chart session first.
    \item Use Alt4Blind to upload the chart and create alternative text for the image provided.
    \item Describe their experience in detail.
\end{enumerate}

During the session, we tracked mouse movements and recorded user interaction behaviors using pen and paper. All participants produced detailed descriptions. P4-P6 especially benefited from the feature that allows copying sentences from various similar charts to craft their descriptions. Expert users appreciated this feature and recommended adding chart captions to further complement the alternative text. P1 suggested increasing access to more than three similar charts, whereas P3 proposed a feature to replace irrelevant images among the selected charts.


\newpage
\section{Discussion and Limitations}
In this section, we shortly discuss our results and the limitations of this work. 
\textbf{Intelligent Features:} Our tool has demonstrated efficiency in enabling both inexperienced and experienced users to author high-quality alt-texts. However, it could further benefit from additional intelligent functionalities, such as LLM-based feedback and descriptions, which are currently under development.\\
\textbf{Captions:} While the current implementation assists users in creating alt-text, we believe that integrating captions into the tool is essential, as captions and alt-texts often complement each other\\
\textbf{Interactions:} Current UI does not allow users to have control over the similar chart section. Future iterations should offer users control, allowing them to omit, replace, or view more charts.\\
\textbf{User Study:} Our initial investigations, though encouraging, were conducted with a limited number of users and lacked comprehensive control measures. Future studies should engage a larger and more diverse group of participants and should involve people with blindness and visual impairments. 
\section{Conclusion}
We presented an online, open-source tool to enhance alt-text writing by using AI image retrieval, making it more engaging. Our shared dataset invites further development by the NLP and Computer Vision communities to advance chart summarization tasks with a focus on people with blindness and visual impairments. Currently, while our work primarily provides examples of alternative texts, it lays the groundwork for future research incorporating Vision-Language models to address accessibility concerns. 



\begin{credits}
\subsubsection{\ackname}
The authors would like to acknowledge the help of P. Venkatesh for his support in developing the UI. We would also like to thank the HoreKa computing cluster at KIT for the computing resources used to conduct this research.

\subsubsection{\discintname}
This research was funded by the European Union’s Horizon $2020$ research and innovation program under the Marie Sklodowska-Curie grant agreements no. $861166$. 

\end{credits}

\newpage
\bibliographystyle{splncs04}
\bibliography{mybib}

\begin{thebibliography}{10}
\providecommand{\url}[1]{\texttt{#1}}
\providecommand{\urlprefix}{URL }
\providecommand{\doi}[1]{https://doi.org/#1}

\bibitem{chintalapati2022dataset}
Chintalapati, S.S., Bragg, J., Wang, L.L.: A dataset of alt texts from hci publications: Analyses and uses towards producing more descriptive alt texts of data visualizations in scientific papers. In: Proceedings of the 24th International ACM SIGACCESS Conference on Computers and Accessibility. pp. 1--12 (2022)

\bibitem{disability2004web}
Commission, D.R.: The web: Access and inclusion for disabled people; a formal investigation. The Stationery Office (2004)

\bibitem{han2023chartllama}
Han, Y., Zhang, C., Chen, X., Yang, X., Wang, Z., Yu, G., Fu, B., Zhang, H.: Chartllama: A multimodal llm for chart understanding and generation (2023)

\bibitem{jeong2023wataa}
Jeong, H., Chun, M., Lee, H., Oh, S.Y., Jung, H.: Wataa: Web alternative text authoring assistant for improving web content accessibility. In: Companion Proceedings of the 28th International Conference on Intelligent User Interfaces. pp. 41--45 (2023)

\bibitem{kantharaj2022chart}
Kantharaj, S., Leong, R.T.K., Lin, X., Masry, A., Thakkar, M., Hoque, E., Joty, S.: Chart-to-text: A large-scale benchmark for chart summarization. arXiv preprint arXiv:2203.06486  (2022)

\bibitem{lundgard2021accessible}
Lundgard, A., Satyanarayan, A.: Accessible visualization via natural language descriptions: A four-level model of semantic content. IEEE transactions on visualization and computer graphics  \textbf{28}(1),  1073--1083 (2021)

\bibitem{mack2021designing}
Mack, K., Cutrell, E., Lee, B., Morris, M.R.: Designing tools for high-quality alt text authoring. In: Proceedings of the 23rd International ACM SIGACCESS Conference on Computers and Accessibility. pp. 1--14 (2021)

\bibitem{meng2024chartassisstant}
Meng, F., Shao, W., Lu, Q., Gao, P., Zhang, K., Qiao, Y., Luo, P.: Chartassisstant: A universal chart multimodal language model via chart-to-table pre-training and multitask instruction tuning (2024)

\bibitem{radford2021learning}
Radford, A., Kim, J.W., Hallacy, C., Ramesh, A., Goh, G., Agarwal, S., Sastry, G., Askell, A., Mishkin, P., Clark, J., et~al.: Learning transferable visual models from natural language supervision. In: International conference on machine learning. pp. 8748--8763. PMLR (2021)

\bibitem{tang2023vistext}
Tang, B.J., Boggust, A., Satyanarayan, A.: Vistext: A benchmark for semantically rich chart captioning. arXiv preprint arXiv:2307.05356  (2023)

\end{thebibliography}
\end{document}